\documentclass[10pt]{article}
\newif\ifblindreview

\usepackage[letterpaper]{geometry}
\usepackage{amta2024}
\usepackage{times}
\usepackage{url}
\usepackage{latexsym}
\usepackage{natbib}
\usepackage{layout}
\usepackage{multicol}
\usepackage{booktabs,array}
\usepackage{float}
\usepackage[hidelinks]{hyperref}
\usepackage[all]{hypcap}
\usepackage{graphicx}
\usepackage{inconsolata}
\usepackage{xcolor}
\hypersetup{
    colorlinks,
    linkcolor={red!50!black},
    citecolor={blue!50!black},
    urlcolor={blue!80!black}
}

\ifblindreview
  \newcommand{\authorinfo}{\author{}} 
\else
  \newcommand{\authorinfo}{
    \author{
            \name{\bf Richard Yue} \hfill \addr{yue.r@northeastern.edu}\\
            \addr{\small Northeastern University, San Jose, CA}
    \AND
            \name{\bf John E. Ortega} \hfill \addr{j.ortega@northeastern.edu}\\
            \addr{\small Institute for Experiential AI, Northeastern University, Boston, MA}
    \AND
           \name{\bf Kenneth Ward Church} \hfill \addr{k.church@northeastern.edu}\\
            \addr{\small Institute for Experiential AI, Northeastern University, Boston, MA}
    }
  }
\fi

\usepackage{todonotes}

\begin{document}

\amtaHeader{x}{x}{xxx-xxx}{2015}{45-character paper description goes here}{Author(s) initials and last name go here}
\title{On Translating Technical Terminology: A Translation Workflow for Machine-Translated Acronyms}
\authorinfo

\maketitle
\pagestyle{empty}

\begin{abstract}
\vspace{5pt}
  The typical workflow for a professional translator to translate a document from its source language (SL) to a target language (TL) is not always focused on what many language models in natural language processing (NLP) do -- predict the next word in a series of words. While high-resource languages like English and French are reported to achieve near human parity using common metrics for measurement such as BLEU and COMET, we find that an important step is being missed: the translation of technical terms, specifically acronyms. Some state-of-the art machine translation systems like Google Translate which are publicly available can be erroneous when dealing with acronyms -- as much as 50\% in our findings. This article addresses acronym disambiguation for MT systems by proposing an additional step to the SL--TL (FR--EN) translation workflow where we first offer a new acronym corpus for public consumption and then experiment with a search-based thresholding algorithm that achieves nearly 10\% increase when compared to Google Translate and OpusMT.
\end{abstract}

\begin{multicols}{2}

\section{Introduction}

With the myriad of artificial intelligence tools available for professional translators, it can be hard for translators to select solutions that address their core needs. Ideally, translation approaches based on machine learning techniques should improve translator proficiency and achieve higher overall quality. One such approach focuses on \textit{technical terminology} (TT) where domain-specific terms in the form of acronyms in a source language (SL) must be translated into their target language (TL) counterpart.

TT is considered important to translators as it is one of the main sources of error a professional translator might encounter on a daily basis. The importance of TT is further displayed by the latest machine translation (MT) workshops \citep{semenov-etal-2023-findings,molchanov2021promt,hasler2018neural} that stress the importance of correctly addressing terminology issues—including correctness of technical terms. While modern MT systems do not seem to focus on acronym and term disambiguation\footnote{MT research generally use metrics such as BLEU \citep{papineni2002bleu} or COMET \citep{rei2020comet}.}, workshops like the “Machine Translation using Terminologies” workshop\footnote{\url{https://www.statmt.org/wmt21/terminology-task.html}} \citep{jon2021cuni} clearly state that they focus on both translation accuracy and consistency. Since the dominant metric used (BLEU) for most MT approaches does not center so much on terminology expansion with acronyms and other mechanisms, we present in this article a novel method that hones in specifically on the day-to-day work in terminology that a professional translator may encounter, which has not been addressed by most of the recent literature.

In this article, we present two main novelties that are based on the translation of acronyms: (1) the introduction of a new corpus made publicly available for others to use and (2) a fact-checking step that is used to verify the combination of a technical term and its acronym (long form (LF) and short form (SF)). We do this for several published articles in the TL, which is English. We aim to show that acronym disambiguation can improve term error rate by reducing the risk of default MT models that generally do not have an acronym approach. Our claim is that translators can use this method as a novel verification step in the normal translation pipeline. We also believe that other automated work such as generative MT may be able to include this step as a mechanism for evaluation.

To that end, we present the following sequence. First, we introduce relevant work in Section \ref{sec:related-work}. Second, we describe our motivation and high-level proposal methodology in Section \ref{sec:background}. We then cover the details of our corpus creation in Section \ref{sec:corpus}. Afterwards, we show the results of our SF/LF method in Section \ref{sec:results} and finally we conclude our work in Section \ref{sec:conclusion}.

\section{Related Work}

\label{sec:related-work}

In the WMT 2023 Shared Task on Machine Translation with Terminologies, \cite{semenov-etal-2023-findings} emphasize the effectiveness of incorporating terminology dictionaries and respecting domain-specific terminology constraints. The authors also distinguish terminology incorporation from general MT methods.

\cite{post2019exploration} explore the use of masking to properly isolate and translate specific named entities such as terminology spans. Their findings show that masking solves some problems, but relies entirely on the masks being properly aligned.

\cite{ghazvininejad2023dictionary} propose a method for translating rare words such as technical terminology. The method, called DiPMT, is a prompting technique that provides an LLM with multiple translation choices from a dictionary as well as hints about their meaning for a subset of input words. It outperforms baselines for low resource and out-of-domain MT. The authors also extract bilingual dictionaries from the training data to assist in this process. Doing so allows for fine-grained control over the use of domain-specific terminology.

\cite{anastasopoulos2021evaluation} stress the importance of taking terminology into account in neural MT and propose metrics to measure MT output consistency with regard to domain constraints. \cite{dagan1994termight} propose a system to identify technical terms in a source text as well as their translations. The system uses part-of-speech tagging and word alignment techniques to assist translators during the translation process. \cite{mckeown1996translating} address the issue of translating collocations in a variety of domains.

\cite{grefenstette1999world} offers an example-based method for dealing with terminology problems in translation as well as other NLP tasks. The method proposed uses search to find the most statistically likely translation of an entire noun phrase. \cite{lee2002translation} provide a knowledge-based approach to translation that includes using word-sense disambiguation to semantically derive the meaning of a word before seeking a target translation corresponding to that meaning.

\cite{skadicnvs2013application} demonstrate the use of a cloud-based terminology search system that fully integrates with statistical methods to address the need for domain-specific terms and their integration into neural MT systems. Meanwhile, \cite{bosca2014lightweight} stress the importance of term verification and consistency in the translation process and propose using external terminological databases to assist in fact checking and correcting domain-specific terminology.

\begin{table*}[!ht]
\centering
\begin{tabular}{lll}
\hline
\textbf{Type of Term} & \textbf{Agreement}\\
\hline \hline
\ Long Forms (LFs) & 62.1\% \\
\ Short Forms (SFs) & 54.3\% \\
\hline \hline
\end{tabular}
\caption{Google Translate agreement for long- and short-form acronyms.}
\label{tab:Google-correctness}
\end{table*}

\section{Background and Motivation}
\label{sec:background}
In order to better understand how ineffective acronym disambiguation may be for translators, we investigate the performance of LFs and their SF acronyms within the realm of commercial MT systems. We perform this necessary step in order to confirm our hypothesis that: \textbf{acronym disambiguation in the current state-of-the-art French MT systems is not being addressed properly}. In Table \ref{tab:Google-correctness}, we provide a specific agreement comparison that uses a widely-used commercial MT system -- Google translate\footnote{\url{https://translate.google.com/}}. For both cases (LFs and SFs) agreement is between 54\% and 63\%, giving way to a high amount of room for improvement. We illustrate this with further analysis in Tables \ref{tab:Google-LF-errors} (long forms) and \ref{tab:Google-SF-errors} (short forms).



\begin{table*}[!ht]
\centering
\begin{tabular}{l | l l}
\hline \hline
\textbf{Input French} & \multicolumn{2}{l}{\textbf{Output English}} \\
 & \textbf{Google} & \textbf{Gold}\\
\hline
\begin{tabular}{l}
indice\\
moteur \\ 
\end{tabular}
& 
\begin{tabular}{l}
engine\\
index\\
\end{tabular}
& 
\begin{tabular}{l}
motricity\\
index\\
\end{tabular}
\\ \hline
\begin{tabular}{l}
fréquence\\
cardiaque\\
\end{tabular}
& 
\begin{tabular}{l}
cardiac\\
frequency\\
\end{tabular}
& 
\begin{tabular}{l}
heart\\
rate\\
\end{tabular}
\\ \hline
\begin{tabular}{l}
roue\\
polaire\\
\end{tabular}
& 
\begin{tabular}{l}
polar\\
wheel\\
\end{tabular}
& 
\begin{tabular}{l}
claw\\
pole\\
\end{tabular}
\\ \hline
\hline
\end{tabular}
\caption{Erroneous Google Translate examples on long forms (LFs).}
\label{tab:Google-LF-errors}
\end{table*}

\begin{table*}[!ht]
\centering
\begin{tabular}{l | ll}
\hline \hline
\textbf{Input French} & \multicolumn{2}{l}{\textbf{Output English}} \\
 & \textbf{Google} & \textbf{Gold}\\
\hline 
AOMI & PAAD & PAD \\
DE & DE & EE \\
ICMI & CIMI & CLI \\
\hline \hline
\end{tabular}
\caption{Erroneous Google Translate examples on short forms (SFs).}
\label{tab:Google-SF-errors}
\end{table*}


As a way of mitigating the room for improvement, we propose the following novel method for MT that decomposes translation into four high-level steps by taking into account that Google Translate is more successful on LFs than SFs. For other MT systems, this may not be the case; we focus solely on Google Translate here as the oracle for our experiment.

\begin{enumerate}
     \setlength{\itemsep}{0pt}
    \setlength{\parskip}{0pt}
    \setlength{\parsep}{0pt}
    \item Use Google Translate to translate each LF from French (FR) to English (EN).
    \item Extract the LF from Google Translate’s EN pair output (using a simple split command).
    \item Generate several SF hypotheses using the extracted LF from Step 2.
    \item Use a search technique to verify and evaluate certainty of hypotheses.
\end{enumerate}


To better describe Steps 1 through 4, we provide the following in-depth description. A term such as “acide désoxyribonucléique \textbf{(adn)}” would first be translated in Step 1 from French to English as “deoxyribonucleic acid \textbf{(dna)}”.  We then extract the English LF (deoxyribonucleic acid) and SF (dna) for use in the next steps. Step 3 consists of the use of AB3P\footnote{\url{https://github.com/ncbi-nlp/Ab3P}}
\citep{sohn2008abbreviation,church2021acronyms}, an acronym tool that provides LFs in English created by the United States government and contains acronyms from crawls of PubMed\footnote{\url{https://pubmed.ncbi.nlm.nih.gov/}} and arXiv\footnote{\url{https://arxiv.org}}. If a sufficient number of documents is not found that contain the English LF and SF together, we then generate a list of acronym hypotheses translations from the translated LF. Each hypothesis is generated using a fine-tuned version of the Scibert \citep{beltagy2019scibert} model described in section \ref{sec:experimental-settings}.

Step 4 consists of the verification process, also known as “Fact Checking”. Typically, the translation process for technical terms involves a significant component of researching the meaning of a source language term, identifying multiple target language candidate terms, and finally, proceeding through the n-best list in order and seeking out the use of a chosen term in context in similar target language texts, written by experts in the field in question.\footnote{\url{https://www.technitrad.com/how-to-perform-terminology-research/}} According to \cite{bowker2021machine}, professional translation term verification is done on the basis of observed frequency in a corpus; if enough experts use the selected term in context, it is considered to be valid. Domain expertise from professional translation trade unions such as the ATA\footnote{\url{https://www.atanet.org/growing-your-career/terminology-management-what-you-should-know/}} point to two or three sources being sufficient to substantiate use of a given term. We replicate that process using the search method below.

We implement a Boolean retrieval system that contains acronyms extracted from AB3P output on a crawl of arXiv and Pubmed along with the long forms they map to and source paper ID. If a sufficient number of sources have been found to employ the desired term-acronym pair (in the form \textit{cardiopulmonary resuscitation (CPR)}), term validation is deemed to be successful and the term pair is returned to the user alongside the list of sources for verification. This re-appropriates the term verification method employed by professional translation agencies in the field (and facilitates verification by a reviewer, who may need to fact check term sources at a later stage).

The translation of acronyms is further complicated by non-English languages opting to adopt a better known English acronym alongside a translation of the term. The French translation for ``large language model'' (grand modèle de langue) is condensed using the English acronym ``LLM,'' even though the acronym does not correspond to the first letters of each word. Despite this limitation, our search step allows for the verification of such cases, as the pairing of term and acronym is likely to occur in the literature if they have found consensus in the field. Thus, verification would succeed and the disambiguation step would not be performed. Furthermore, fine tuning on corpora such as Pubmed was foregone due to the non-compositionality of many technical terms; boolean search ensures that the term is verified as a fixed unit.

\begin{table*}[!ht]
\centering
\begin{tabular}{p{5cm} | l}
\hline
\hline
\textbf{Input: LF ([MASK])} & \textbf{Gold SF} \\
\hline
cardiopulmonary resuscitation
([MASK])& CPR  \\ \hline
deoxyribonucleic acid ([MASK]) &  DNA \\ \hline
Organization of the Petroleum Exporting Countries ([MASK]) & OPEC\\ \hline
\hline
\hline
\end{tabular}
\caption{Training data for SF candidate generation.}
\label{tab:training_data} 
\end{table*}
\begin{table*}[!ht]
\centering

\begin{tabular}{l p{3cm} l}
\hline
\hline
\textbf{Baseline} & \textbf{Input} & \textbf{Output} \\
\hline
Identity & ADN & ADN \\ \hline
Reverse & ADN & NDA \\ \hline
Google/Opus & acide désoxyribonucléique (ADN) & DNA \\ \hline
\hline
\end{tabular}
\caption{Examples of our three baseline methods.}
\label{tab:example-of-baselines}
\end{table*}

While an exact match (e.g. ‘RCP’ to ‘CPR’) is the objective of our system, it is important to note that for evaluating the system we distinguish between \textit{agreement} (an exact match) and \textit{verification} (verified by a search) as noted:

\begin{description}
     \setlength{\itemsep}{0pt}
    \setlength{\parskip}{0pt}
    \setlength{\parsep}{0pt}
   \item[Agreement] -- The candidate SF is an exact match with the gold SF.
   \item[Verification] -- The candidate SF was found near the LF in at least two published papers in the target language (English).
\end{description}








\section{Experimental Settings}

\label{sec:experimental-settings}

\subsection{Translation Models}

For \textbf{Google Translate}, experiments were performed using the Google API\footnote{\url{https://cloud.google.com/translate}} as available to the public on October 14, 2023. For \textbf{Opus MT}, the vanilla model was used without any fine tuning. The French-to-English language variant from Hugging Face\footnote{\url{https://huggingface.co/Helsinki-NLP/opus-mt-fr-en}} was downloaded for this purpose.

\subsection{Baselines}

We compare the inclusion of our method against several baselines that are executed with and without our proposed step. Our experiments are performed on the acronym corpus that we created and allow for public consumption. Our first set of experiments focuses on three main baseline approaches found in Table \ref{tab:example-of-baselines} that we call: (1) \textit{Identity}, (2) \textit{Reverse}, and (3) \textit{Google/Opus}\footnote{We use the OpusMT system for an extra comparison \url{https://huggingface.co/Helsinki-NLP/opus-mt-fr-en}}. The Identity baseline is the most straightforward experiment which is when the English SF output is \textbf{equal to} the French SF input (e.g. ADN in French is equal to ADN in English). The Reverse baseline is when the English SF output is the \textbf{reverse} of the French SF input (e.g. ADN in French is equal to NDA in English). The Google/Opus baseline takes the LF and SF in French and outputs an SF in English.

\subsection{Hypothesis Generation}

For the disambiguation of acronyms, we use a SciBERT \citep{beltagy2019scibert} model that is fine-tuned on 1.8M term-acronym pairs in the target language (English) with these parameters: Adam as the optimizer, an initial learning rate of 2e-5, 1,000 warmup steps, and a weight decay of 0.01.  We use data downloaded from arXiv\footnote{\url{https://info.arxiv.org/help/bulk_data/index.html}} and then processed by AB3P for fine-tuning as shown in Table \ref{tab:training_data}. The final model accepts input in the form: “LF ([MASK])” and outputs an n-best list of SF candidates.

\subsection{Acronym Corpus}
\label{sec:corpus}

A new test set\footnote{\url{https://github.com/rtotheich/acronym_corpus/tree/main}} (called the \textbf{acronym corpus} here) has been created for evaluating machine translation systems on acronyms. 
The test set consists of 437 LF-SF pairs obtained from a corpus of 13,500 abstracts crawled from HAL\footnote{\url{https://theses.hal.science/?lang=en}}, a repository of French 
academic papers, many of which are from medicine and science. The pairings contain an LF and SF for each term in both French (source) and English (target). Examples were selected such that no offensive content or personal information was to be included.

The HAL repository provides abstracts in both French and English. These abstracts contain many technical terms. An example of an abstract is “[...] 42/194 patients (21\%) did not want \textbf{cardiopulmonary resuscitation (CPR)} and 15/36 (41\%) did not prefer intensive care unit (ICU) admission [...].” When the abstract introduces an acronym, the gold labels in the test set specify the long form (LF) and the short form (SF) in both French and English.  An example of the acronym translation task is to input a French LF such as \textbf{réanimation cardiopulmonaire} and its corresponding SF, in this case \textbf{RCP}. The output should be the correct translation of the SF: \textbf{CPR}.

\begin{table*}[!ht]
\centering
\begin{tabular}{lll}
\hline
\hline
\textbf{Method} & \textbf{Agreement} & \textbf{Verified} \\
\hline
Identity Baseline & 21.5\% & 0.06\% \\
Reverse Baseline & 28.5\% & 14.6\% \\
Opus Baseline & 34\% & 14.9\% \\
Google Baseline & 54.3\% & 29.2\% \\
Gold Labels & 100\% & 42\% \\
\hline
\hline
\end{tabular}
\caption{Agreement and verification for the baseline experiments on the 
\textit{Acronym Corpus}.}
\label{tab:results} 
\end{table*}
\begin{table*}[!ht]
\centering
\begin{tabular}{lll}
\hline
\hline
\textbf{Method} & \textbf{Precision} & \textbf{Recall} \\
\hline
Identity Baseline & 0.28 & 0.06 \\
Reverse Baseline & 0.51 & 0.15 \\
Opus Baseline & 0.43 & 0.15 \\
Google Baseline & 0.54 & 0.29 \\
Gold Labels & 0.42 & 0.42 \\
Proposed (Opus) &  0.75 & 0.33 \\
Proposed (Google) &  \textbf{0.68} & \textbf{0.43} \\
\hline
\hline
\end{tabular}
\caption{Precision and recall comparisons for all experimental systems.}
\label{tab:precision_recall} 
\end{table*}

\section{Results}
\label{sec:results}





We compare the baselines first in Table \ref{tab:results}. We provide both agreement and verification for consistency purposes, which show that verification is generally much lower than agreement for all systems.


When compared, our proposed technique, which includes search and verification, achieves 9.9\% improvement (43.9\%) for agreement and 17.8\% improvement (32.7\%) for verification compared to the baseline when using the OpusMT system. Google translate scores are also markedly higher, with 8.3\% improvement (\textbf{62.6\%}) and 13.6\% (\textbf{42.8\%}), respectively. It is clear that through the use of our proposed system, the acronym resolution is much higher for both agreement and verification.



Additionally, we illustrate the comparisons in more detail from a precision and recall perspective in Table \ref{tab:precision_recall} for all experimental systems. Our experiments show that through the use of our proposed step which uses agreement and verification, professional translators that use the Annotated Corpus will have more success using our system. Precision is presented here as the portion of agreed terms that are verified and recall as the portion of verified terms.

\section{Conclusion}

\label{sec:conclusion}

Professional translators must be well versed in the source and target languages that they are translating. Translating technical terminology can be so important that it has been compared to the job of a terminologist \citep{cabre2010terminology}. Quality translations will take into account several units of measurement such as fluency, adequacy, and more. However, it has been the case in the past that, more often than not, terminology, specifically the translation of acronyms, is not included as a major improvement to a translator’s pipeline. Domain-specific standards \citep{ghenctulescu2015importance}, nonetheless, have been set such that verification of terminology like acronyms is considered an important step in translation.

Translators and AI practitioners could benefit highly from the use of a system like the one presented in this article. We believe that our corpus and findings provide sufficient evidence and materials to reproduce a benefit to warrant future work on the topic.



\section{Limitations}

The results of applying our method may not transfer to languages
that are very different from English in orthography (e.g., Chinese, Japanese) and/or morphology. The working languages of the authors being French and English, hand curating a corpus was limited to these only. Our solution also may not scale to longer texts; the method is based on working with term-acronym pairs and working on a full text would require a pre-processing step to identify term pairs as well as inference time for each acronym. Training a model for this task also requires access to GPU resources.

\section{Ethics Statement}

In line with the concept of professional translator ethics presented by \cite{lambert2020professional}, it is of paramount importance to guard against translations that “represent their source texts in unfair ways.” This refers to unfaithful translations that do not correctly transfer the true meaning in the source language, a prime example being incorrect or unverifiable terminology. Our system upholds this doctrine of translation ethics and adheres to ethics policies outlined by the translation community.
\newpage

\begin{small}
\bibliographystyle{apalike}
\bibliography{amta2024}
\end{small}

\end{multicols}

\end{document}